%% file: ms.tex
\documentclass[10pt,letterpaper]{article}

\usepackage{cogsci}

\cogscifinalcopy % Uncomment this line for the final submission 
\usepackage{amsmath}
\usepackage{color}
\usepackage{soul}
\usepackage{pslatex}
\usepackage[natbibapa]{apacite}
\usepackage{float} % Roger Levy added this and changed figure/table
\usepackage{graphicx}
\usepackage[compact]{titlesec} %space around section headings
\usepackage{todonotes}

\titlespacing{\section}{0pt}{*1}{*1}
\titlespacing{\subsection}{0pt}{*1}{*1}
\titlespacing{\subsubsection}{0pt}{*0.1}{*1}

\title{Explaining intuitive difficulty judgments by modeling physical effort and risk}
 
\author{
{\bf Ilker Yildirim*$^{1,2}$} {\normalfont (ilkery@mit.edu),} Basil Saeed*$^1$ {\normalfont (bsaeed@mit.edu),}\\
{\bf Grace Bennett-Pierre$^{3}$} (gbp@stanford.edu), {\bf Tobias Gerstenberg$^3$} (gerstenberg@stanford.edu),\\
{\bf Joshua B. Tenenbaum+$^1$} (jbt@mit.edu), {\bf Hyowon Gweon+$^3$} (hyo@stanford.edu)\\[0.5em]
$^1$ Department of Brain and Cognitive Sciences, MIT $^2$ Department of Psychology, Yale University\\
$^3$ Department of Psychology, Stanford University\\[0.2em]
* indicates co-first authors; + indicates co-senior authors
}

\begin{document}

\maketitle

\begin{abstract}
The ability to estimate task difficulty is critical for many real-world decisions such as setting appropriate goals for ourselves or appreciating others' accomplishments. Here we give a computational account of how humans judge the difficulty of a range of physical construction tasks (e.g., moving 10 loose blocks from their initial configuration to their target configuration, such as a vertical tower) by quantifying two key factors that influence construction difficulty: physical effort and physical risk. Physical effort captures the minimal work needed to transport all objects to their final positions, and is computed using a hybrid task-and-motion planner. Physical risk corresponds to stability of the structure, and is computed using noisy physics simulations to capture the costs for precision (e.g., attention, coordination, fine motor movements) required for success. We show that the full effort-risk model captures human estimates of difficulty and construction time better than either component alone.

\end{abstract}

\input{intro.tex}
\input{experiments.tex}

\input{model.tex}

\input{results.tex}

\input{discussion.tex}

\section{Acknowledgments}
This work was supported by an NSF STC award CCF-1231216 (Center for Brains, Minds \& Machines), an ONR grant N00014-13-1-0333, a grant from Mitsubishi MELCO (JBT); and a McDonnell Scholars Award (HG).

\vspace{-0.1cm}

\bibliographystyle{apacite}

\bibliography{TobiPapers}

\end{document}

%% file: intro.tex
\section{Introduction}

Suppose two people are playing with blocks. Sally stacks 10 blocks to build a vertical tower (Figure~\ref{fig:stimuli}, from 7-H-A to 7-H-B), and Anne puts 10 blocks side by side to form a horizontal line (Figure~\ref{fig:stimuli}, from 7-E-A to 7-E-B). Which one is harder to make? Even though both Sally and Anne worked with the same number of blocks, we have the intuition that Sally's tower is harder to make than Anne's horizontal line. When prompted to explain why, you might note that Sally's tower is more likely to fall, so she might need to re-stack the blocks, or be more careful than Anne when placing the blocks so they don't fall. 

Intuitive judgments of task difficulty are ubiquitous in our actions and decisions \citep{carroll1981ability}. While as adults, we rarely have to think seriously about the difficulty of building block structures, estimating task difficulty is critical for efficiently navigating the physical and social world; for instance, we often think about how difficult it would be to climb a new route at the climbing gym, follow an unfamiliar recipe for dinner, or finish a conference paper before the deadline \citep{gweon2017}. 

Sometimes, our prior experience can give us a vague sense for the difficulty of similar tasks (e.g., ``making oatmeal is not that hard''). However, judgments of difficulty often have to be made before we actually engage in a task, and estimating task difficulty is particularly important for tasks we've never completed before. For novel tasks, we might want to first evaluate the probability of success to even decide whether to try that task at all, set more reasonable goals, or decide whether additional help is necessary \citep{gweon2011rationally}. In these cases, one must represent the current state of the world, the desired goal state, and make a plan for how to get from the initial state to the final goal state.  

The explanations that we generate can sometimes provide a useful glimpse into the underlying computations. As in the example above, to explain why Sally's tower is harder to build than Anne's line, we may allude to the physics behind the building process (e.g., stability of the structures and the probability of falling over) and the nature of the actions required to complete the task. Yet, it remains unclear whether these explanations reflect the actual inferential processes that give rise to people's judgments. What representations and inferential processes underlie our intuitive judgments of task difficulty?

\subsection{Estimating difficulty}

Difficulty is an abstract, unit-less construct that cannot be measured on a single, absolute scale. Difficulty incorporates both task-specific properties (e.g., Task~A can be more difficult than Task~B) and agent-specific attributes (e.g., a task can be more difficult for Tom than for Suzy depending on their competence). Difficulty can refer to both physical effort (e.g., force, work, time, fine motor control) as well as mental effort (e.g., attention, cognitive control, inhibition, or even requisite level of skill or intelligence). Even in making seemingly simple inferences such as judging Sally's tower as harder to make than Anne's line, people's explanations often refer to both types of effort: Sally's actions need to be more precise (physical effort) and she needs a higher level of concentration to correctly execute her plan (psychological effort). Unlike the height or weight of the structures that have clear ground-truth values that can be measured in objective units (e.g., 10 inches tall, 5 ounces in weight), the difficulty of building a structure, while certainly related to its height or weight, cannot be easily expressed as a single point estimate. 

\begin{figure*}[ht!]
\begin{center}
\setlength{\fboxsep}{0pt}
\setlength{\fboxrule}{0.5pt}
\fbox{\includegraphics[width=\textwidth]{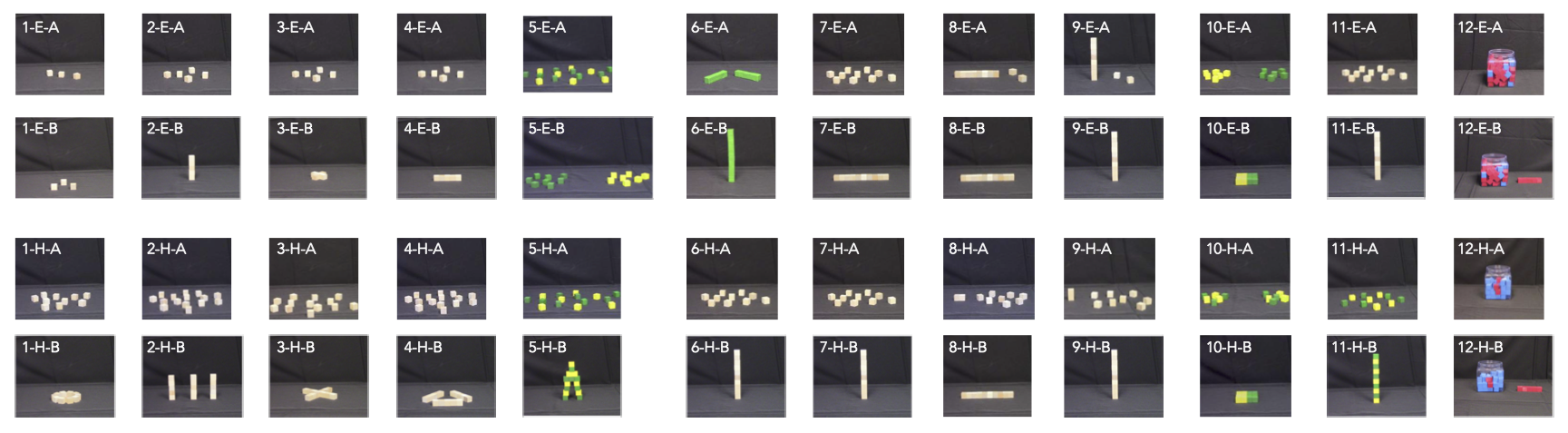}}
\end{center}
\vspace{-0.5cm}
\caption{Experimental stimuli comprise 24 trials, each with an initial state (A) and final state (B). Trials are divided into 12 matched pairs, with one task that is intuitively easier  and another harder, marked E, H respectively. (Note that E and H are relative designations within a given pair of trials. A similar task may be easy in one pair and hard in another, e.g. pairs 7, 11.)} 
\label{fig:stimuli}
\vspace{-0.5cm}
\end{figure*}

Nevertheless, there is reasonable consistency and systematicity in human observers' estimates of task difficulty. Recent work suggests that adults can provide difficulty estimates that are tightly correlated with the time it takes to build a structure, and that even preschool-aged children are quite accurate at judging the relative difficulty of two block structures similar to those shown in Figure~\ref{fig:stimuli} \citep{gweon2017}. Critically, these judgments and estimates were made from simply looking at still photographs of the structures, rather than having actually engaged in the building process. This suggests that even though difficulty may not have an objective unit that cannot be directly measured, people use proxy measures like time or amount of work to communicate their intuitions in some measurable format (e.g., this structure might take 15 seconds to complete), and even use these estimates to generate ad-hoc scales that allow them to express difficulty in relative terms (e.g., Task~X is more difficult than Task~Y). However, the cognitive mechanisms that support these computations remain poorly understood. 

\subsection{Modeling difficulty}

Here we give a computational account of these behaviors by drawing on recent work in robotics and cognitive science. We assume that the observer estimates both the \emph{physical effort} required to carry out an efficient plan as well as the \emph{physical risk} arising from the potential instability of the target construction. We model physical effort as the total kinetic energy required to intervene on the world to change the current state to the desired final state; we do this by using a hybrid task-and-motion planner that first finds a symbolic action plan (e.g., pick, place, hold), and then translates this symbolic plan into continuous motion trajectories of the agent's actions. We model physical risk using noisy physics simulations to capture aspects of the construction process that go beyond just transporting blocks from their initial to their final positions. In the face of risk (i.e. potential physical instability), our agent increases its motor precision to create a more stable structure.%robustly build the the target structure. 
There are a number of ways in which an agent might increase precision; to minimize noise, an agent might engage in fine motor control and coordination, or deploy cognitive control or attention. All these efforts to ensure appropriate level of precision is reflected in physical risk, a probabilistic estimate of the stability of the final target state. 
Recent work on intuitive physical reasoning suggests that humans have an  ``intuitive physics engine'' (IPE) that allows them to predict future physical states of the world by engaging in mental simulation \citep{battaglia2013simulation,gerstenberg2017faulty}. %[describe briefly?]. 
However, the focus of this prior work was primarily on explaining how people predict the future states of the world in the absence of any agent interventions, such as judging whether a block tower will fall given the weight of its components or in which direction the tower is likely to fall. Yet, an internal physics engine is also critical for action planning and object manipulation. Indeed, most advanced object manipulation systems in robotics combine both agent models and physics \citep{toussaint2015logic,todorov2018goal,toussaint2018differentiable}. 

Imagine an agent in an ``initial state'' where 10 loose blocks lie scattered on the floor (Fig. 1, 7-H-A). The agent could perform the first intervention by picking up a block, moving it across the space, and placing it at a particular position. The next action could involve picking up another one, and placing it on top of the first one. By combining the IPE with an intuitive understanding of goal-directed actions of an agent, human observers can simulate forward the process of building a block structure. Human observers can also reason backwards from a representation of the final completed task to estimate the number of actions, or the time that it took, to build the tower  \citep{gweon2017}. Recent work suggests that even children engage in physical problem solving in ways that are similar to adults \citep{cortesa2017characterizing,cortesa2018constraints}. One intriguing possibility is that the kinds of proxy measures that human subjects use intuitively (e.g., time to completion, etc.) are generated from such process of physical simulation conditioned on agent action. 

Here we evaluate this hypothesis through computational modeling. The process of building such block structures lends itself well to precise formalization. The physical principles involved in these block-building tasks are captured in Newtonian mechanics (e.g., gravity, forces including support relations, motion). Existing physics engines can simulate these principles efficiently and realistically \citep{coumans2010bullet}, and are successfully utilized to explain aspects of human physical scene understanding as noted above \citep{battaglia2013simulation,gerstenberg2017faulty}. Here, building on this and other prior work \citep{yildirim2017physical}, we focus on how IPE might be deployed in our everyday reasoning about actions and their consequences. More specifically, we use IPE model to estimate the physical risk (stability) of a given plan; we argue that people's intuitive estimates of task difficulty are rooted in people's abilities to engage in action planning by simulating an agent's goal-directed physical manipulation of objects. 

%% file: experiments.tex
\section{Experiments: Estimated and actual difficulty}

We collected data from human adults to characterize the actual difficulty (measured in building time) of the block structures as well as their estimates of expected building time (in seconds) and estimates of difficulty (on an arbitrary scale). 

\subsubsection{Participants}
The data came from three separate tasks. 1) \textit{Difficulty Estimation task}: $n = 59$ (20 females), $M_\text{age}=31.62~(\text{SD} = 8.65)$; 2) \textit{Time Estimation task}: $n = 60$ (23 females), $M_\text{age} = 32.73~ (8.72)$; and 3) \textit{Build Task}: $n = 20$, $M_\text{age}= 18.75~(2.20)$. The Difficulty Estimation and Time Estimation tasks were conducted on Amazon MTurk. Two participants were excluded for providing identical responses on all trials. 
The Build task was conducted in the lab (these data were originally presented in \citealp{gweon2017}). 

\subsubsection{Materials}
Photographs of 24 sets of structures made of 1-inch wooden blocks were used (12 pairs, 1 easy and 1 hard structure). There were two photographs for each structure: ``initial state'' (for example, scattered blocks) and a ``final state'' (for example, a 10 block vertical tower). On the screen they were labeled ``Start'' and ``Finish'' respectively. The order of presentation was randomized across 12 pairs and counterbalanced within a each pair. We also used one particularly ``hard'' and one particularly ``easy'' structure to help anchor their subsequent judgments it the beginning of the task. 

\subsubsection{Procedure}
Depending on the task, participants used a sliding scale to answer one of these questions for every trial: ``How difficult would it be to do this?'' (\textit{Difficulty Estimation Task}), ``How long would it take to make this?'' (\textit{Time Estimation Task}). In order to account for differences in scale use between participants, we z-scored each participant's responses. For the \textit{Build Task} (in-lab), the experimenter laid out blocks in front of the participant in the same configuration as the initial photo on each trial, and asked the participant to create the structure shown in the final state photo. We recorded how long the subject spent building the block structure from start to finish using a key press.

%% file: model.tex
\vspace{-0.4\baselineskip}
\section{Model}

\begin{figure*}
\begin{center}
\includegraphics[width=0.95\textwidth]{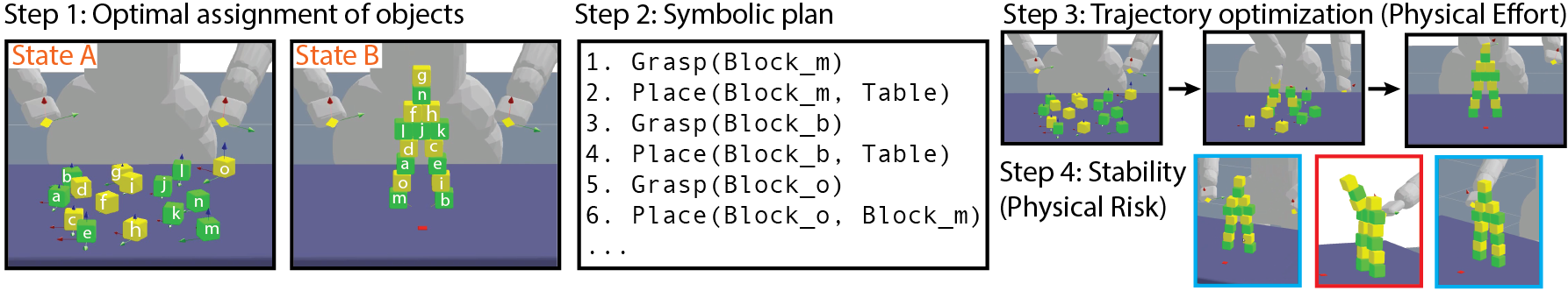}
\end{center}
\vspace{-0.5cm}
\caption{Overview of the modeling approach. In step 1, the model chooses an assignment of objects from state A to B minimizing the total Euclidean distance between initial and final 3D positions. In Step 2, the model finds a symbolic plan that respects the support graph of State B. Two action primitives are sufficient to solve all of our trials, \texttt{Grasp(Obj)} and \texttt{Place(Obj1, Obj2)}. In Step 3, we perform trajectory optimization to find a full 3D path for each block to be transported from an initial to a final position. Finally, in Step 4, we estimate stability (we define physical risk is as 1 - stability) by introducing random but small amounts of perturbations to State B. Blue frames show stable runs, the red frame is unstable.} 
\vspace{-0.5cm}
\label{fig:model}
\end{figure*}

The model consists of two components. The first component evaluates the amount of physical effort that would be required to change an initial spread of blocks on flat surface (Initial State, or State ``A'') to a particular target configuration of blocks (Final State, or State ``B''; see Fig.\ref{fig:model}). We define physical effort as the total kinetic energy required to transport all blocks to change the current state A to B. We derive this quantity using a hybrid task-and-motion planner. This planner is based on the logic-geometric programming framework and consists of an initial stage of symbolic action sequence planning to go from A to B followed by trajectory optimization to fulfill the symbolic plan with respect to a simulated humanoid robot. The planner optimizes for a geometric cost term consisting of the sum of squared accelerations. 

The second component of the model captures the physical risk -- due to potential instabilities -- involved in transitioning from State A to B. For humans, these aspects might include how much attention and motor precision is required to keep the structure from falling during its intermediate stages. We approximate these various facets of physical risk in a probabilistic estimate of the stability of the target state B. We define physical risk as how likely B is to fall, which we estimate by aggregating the outcomes of multiple forward physics simulations each starting from a configuration where the 3-D position of each block in State B is perturbed with a small random noise \citep[cf.][]{battaglia2013simulation}. 

\subsection{Modeling physical effort}
We model physical effort using a hybrid task-and-motion planner -- specifically, the logic-geometric programming solver \citep{toussaint2015logic}. Inputs to this planner are symbolic representations of State A and State B. For each object (i.e. a single block), the input representation specifies its dimensions, position, and support relations (the list of other objects supporting this object). Support relations across all objects can be described in a simple directed graph, which we refer to as a ``support graph'', with nodes indicating objects and edges indicating support relations.

Assignment of identities to objects within each symbolic representation is arbitrary and identities across States A and B need to be matched carefully. To illustrate, consider 11-E-A and 11-E-B in Fig.\ref{fig:stimuli}. All blocks appear identical and there is no reason subjects or the model should assume a particular assignment based on visual appearances of blocks. Instead, in the model, we choose a one-to-one assignment of blocks in State A to blocks in State B that minimizes the total Euclidean distance between the initial and final positions of all blocks while taking into account the blocks' colors (e.g., a green block in 11-H-A cannot be assigned to a yellow block in 11-H-B). To find the mapping with a minimal total Euclidean distance we use the Hungarian Algorithm \citep{kuhn1955hungarian}.

Now given the representations of State A and State B and the optimal mapping of objects from one state to the other, the planner proceeds in two stages: 1) a \textit{symbolic planner} and 2) a \textit{geometric planner}.

\subsubsection{Symbolic planning}

In the first stage, a task planner generates a sequence of abstract actions to transform State A to State B. In our setting, this is equivalent to finding an ordering of blocks by which State B will be constructed from blocks in State A. We find this ordering by exploiting the support graph of state B. We find a sequence of actions that builds state B from the ground up, making sure that each block to be placed has its support already in place. Fig.\ref{fig:model} shows one such assignment of blocks for an example state pair and the resulting symbolic plan by respecting the ordering arising from the support graph of state B. For every block that needs to be moved, we use two action primitives (implementing \textsc{grasp} and then \textsc{place} actions with just a single hand of the robot). The length of the plan ranged from 4 to 30 actions across the trials depending on the number of blocks that were needed to be transported.

This symbolic plan does not take into account the geometry of the shapes, or their rotations in space, and how they would be transported by an efficient agent. Instead, it treats each object as a point object that is supported by another point object. The only relationships this planner respects are the support relations between objects. The distance between blocks is implicitly taken into account due to the optimal assignment of blocks as pairs prior to symbolic task planning.\footnote{Note we are separately handling distance and support relation. Finding distance-minimizing pairs with respect to the topological order in the support graph might yield more efficient possible solutions in some cases.}

\subsubsection{Geometric planning}

The second stage of the planner, the geometric planner computes a full path in 3D for all objects through trajectory optimization. The symbolic plan as well as representations of State A and B are fed to the geometric planner, which uses the geometric information in the initial and final state to create a set of constrained optimization problems. The geometric planner finds the optimal movements that the simulated humanoid robot must take to move the blocks according to the sequence of moves in the symbolic plan. Following \cite{yildirim2017physical}, the planner minimizes squared acceleration of the joints of the simulated humanoid robot using the KOMO (k-order motion-optimization) package introduced in \citep{toussaint2015logic}.

%\subsection{Estimating physical effort}

\subsubsection{Measuring physical effort}

Physical effort is measured by the kinetic energy needed to execute the resulting full plan. Kinetic energy, $\frac12mv^2$, requires knowing the velocity by which the object is moved and its mass value for each step of the symbolic plan. Velocity values are calculated from the geometric planner by using the velocity with which the blocks are being moved for each step of the symbolic plan while the solution is being executed. Note that these velocity values result from the optimization procedure that minimizes the sum of squared accelerations. We assume that the blocks are of equal size and equal mass and assigned a unit mass to a unit block. We obtain the total kinetic energy required to execute a plan for a given pair of states A and B by summing up the energy incurred by each step of the symbolic plan. 

We used $M=30$ 
% ($M=30$ in our simulations) 
random samples to represent uncertainty in the initial configurations that arises from the exact positions of scattered blocks (blocks that are not part of a clear structure; or more specifically blocks that are laying on the table without supporting any other block, e.g., all blocks in Fig. 1 5-E-A).\footnote{We did not systematically evaluate how the value of $M$ would impact our results. However, we noticed that the variation across our $30$ samples were not high to justify a bigger $M$, and it is possible that even smaller values of $M$ will produce very similar results as reported here.} This randomness in the initial configurations creates variation from simulation to simulation in how the blocks are assigned between A and B, and what symbolic and geometric solutions are generated. For a given pair A and B, we calculate the total kinetic energy for each randomly drawn A, and average across all 30 samples.

One trial (pair 12E and 12H in which blocks are placed in a bucket) needed additional assumptions to solve. The planner is incapable of searching for specific blocks from a mix of blocks in the box, or calculating paths that avoid passing blocks through the box. Thus, we handled the initial configurations in the following way: We lined up the 20 blocks in a pre-specified order (a $10 \times 2$ array) and randomly chose which 5 should be colored blue for trial 12-E and which should be colored red for trial 12-H \citep[cf.][]{jara-ettinger2018sensitiv}. Like other trials, we solve and obtain the total kinetic energy for each 30 such random initial configurations, and then average across to obtain an estimate of physical effort.

\subsection{Modeling physical risk}

We define the physical risk associated with a trial by how likely its final configuration B is to fall. We estimate this physical risk following \cite{battaglia2013simulation} by approximating a distribution over how robust the final configuration of blocks B is to small perturbations of the blocks' positions. We ran $N=100$ simulations and recorded whether or not the final configuration remained stable after noise was applied. For each sample, we modified B by randomly perturbing the position of each block in while ensuring that there is no overlap between blocks (see Fig.~\ref{fig:model} for example perturbed configurations). The random perturbations for each block were drawn i.i.d. from a Gaussian distribution with 0 mean and standard deviation $\sigma$. Each perturbed configuration was simulated forward using a physics engine (we used PhysX physics engine in our simulations) either until all motion stopped or until a given period of time elapsed\footnote{To determine whether a tower fell, we simulated it starting from its initial configuration for 4 seconds. If the difference between the final and initial positions of any of the blocks is above a set threshold, we considered that tower to have fallen.}. We then estimated physical risk of the target configuration B to be the fraction of simulations where the tower fell relative to all $N$ simulations. We fitted $\sigma$ to the data using a simple grid search exploring values between $0.05d$ to $0.1d$ with increments of $0.005d$ where $d$ is the length of the the block along the axis it is being perturbed (see beginning of the Results section for the details of the fitting procedure).

\begin{figure}[t]
\begin{center}
\includegraphics[width=\columnwidth]{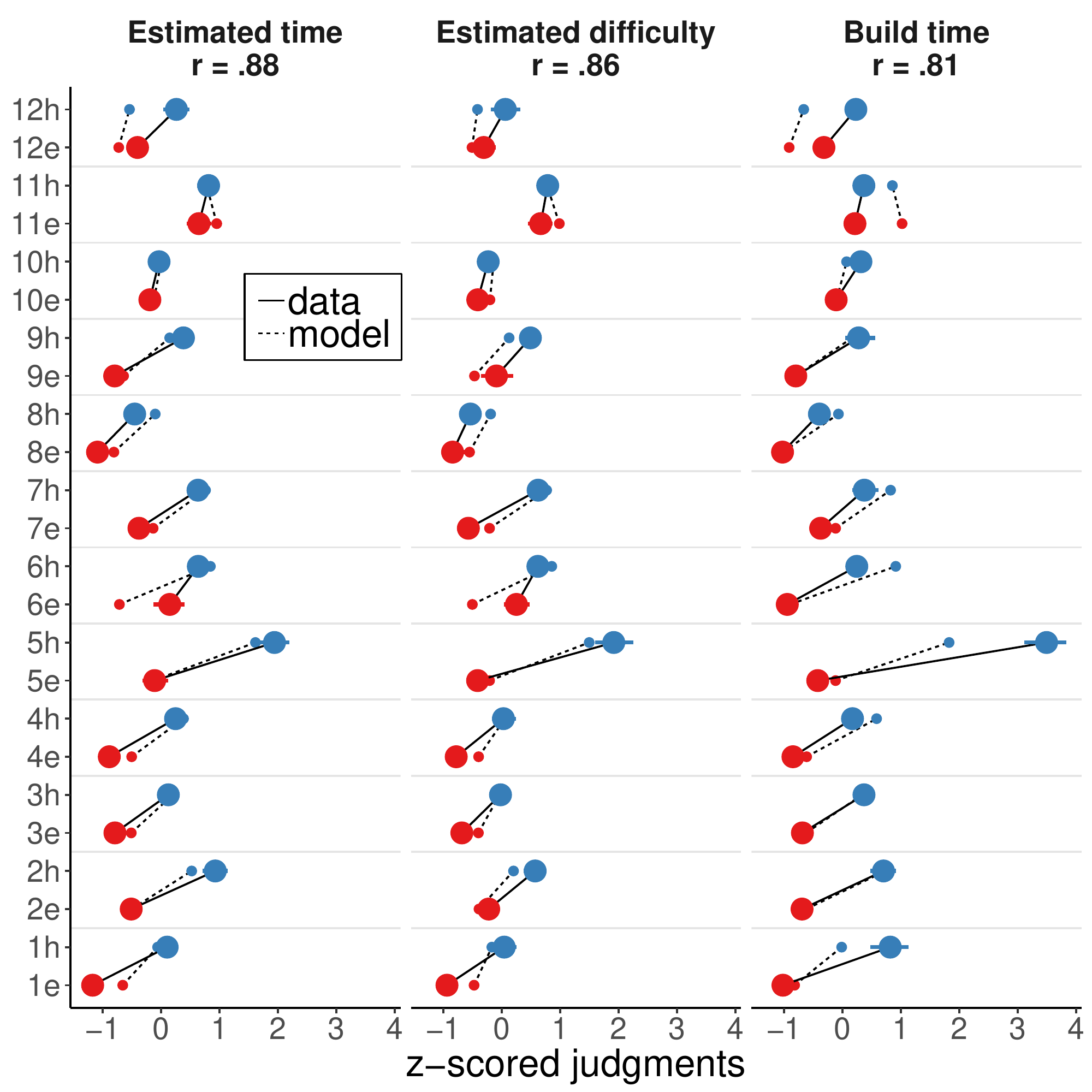}
\vspace{-1cm}
\end{center}
\caption{Mean z-scored human judgments (large points connected by solid lines) and fitted model results (small points connected by dotted lines) with $\text{sigma}=0.065$. The model closely tracks human judgments in each task. It also captures the easy vs. hard distinction in all pairs of easy (red) vs. hard (blue) trials (e.g., 1e vs 1h) except for 11e vs. 11h. Error bars show 95\% bootstrapped confidence intervals.} 
\label{fig:bestfit}
\vspace{-0.2cm}
\end{figure}
% \vspace{-0.5cm}

\subsection{The full model and lesioned models}

We denote the estimated physical effort as $E_i$ and the estimated physical risk as $R_i$ where $i$ indexes the trial number across the 24 trials in our experiments (Fig. \ref{fig:stimuli}). Note that units of $E_i$ are in joules and non-negative, whereas $R_i$ is a probability between $0$ and $1$. Our full model partitions $E_i$ into two factors based on $R_i$ and fits a different coefficient for its risk portion $E_i \times R_i$ and its risk-free portion $E_i \times (1-R_i)$ based on the average human responses:
\vspace{-0.25cm}
\begin{equation}
y_i = \beta_0 + \beta_1 E_i (1-R_i) + \beta_2 E_i R_i
\label{eq:regression}
\end{equation}
where $y_i$ denotes average human response for trial $i$ and $\beta_0, \beta_1,$ and $\beta_2$ are coefficients. 

We compare the full model with lesioned models that only consider the physical effort, or the physical risk. The \textit{physical effort only} model was fit using $y_i = \beta_0 + \beta_1 E_i$. The \textit{physical risk only} model was fit $y_i = \beta_0 + \beta_1 R_i$. Since we evaluate the \textit{physical risk only} model merely with respect to how well it correlates with the data (see Fig.~\ref{fig:allmodels}), fitting a slope and intercept does not affect the results.

%% file: results.tex
\section{Results}

\begin{figure}[t]
\begin{center}
\includegraphics[width=\linewidth]{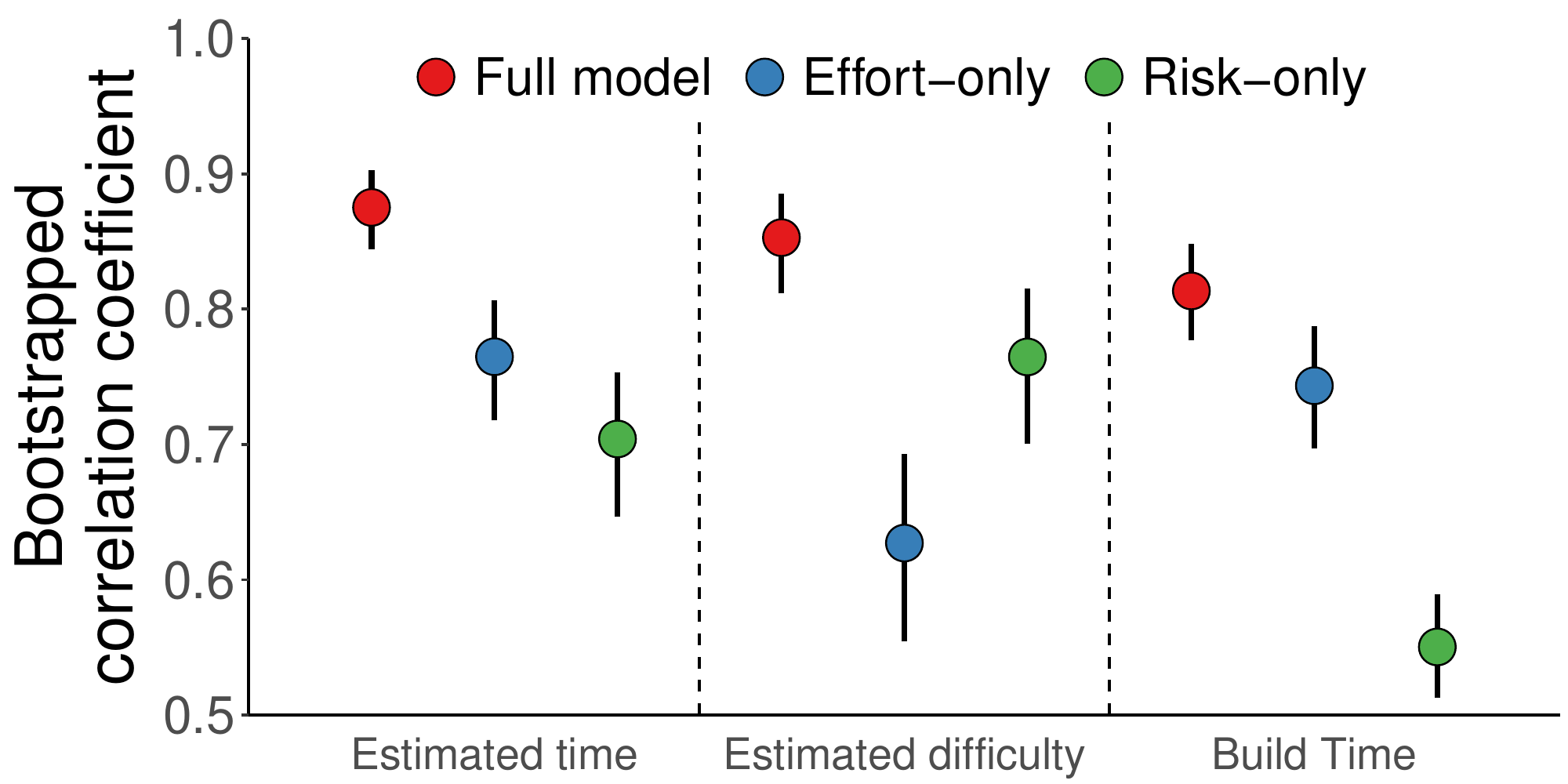}
\end{center}
\vspace{-0.4cm}
\caption{Comparisons of the full model, effort-only model, and risk-only model. Filled circles show median correlation values based on 1000 bootstrapped samples. Error bars show the 2.5 and 97.5 percentile.} 
\label{fig:allmodels}
\vspace{-0.5cm}
\end{figure}

In the full model, we estimated the noise parameter $\sigma$ for the physical risk estimation component of the model by fitting the linear regression in Equation~1 for each dataset individually, and then exploring the root mean squared error (RMSE) between model predictions and behavior. 
We set $sigma$ to $0.065$ (RMSE was higher for lower noise levels, and didn't change much for a range of levels). 
The full model which combines effort and physical risk better captures the data than the effort-only model (Figure~\ref{fig:allmodels}).

Figure~\ref{fig:bestfit} shows the best fitting full model results against the data at the trial-level granularity for each task. The full model's (combined physical effort and risk) predictions closely tracked the data and are correlated highly with each of the three tasks. Not only did the full model capture the overall distribution of responses, it also recapitulated the easy versus hard distinction in most trial pairs. The model produces the correct rank order (assigning a smaller cost to the easy pair than the hard pair) in a11 of the pairs. The only exception is the pair 11E and 11H, where humans did not show any difference across all three datasets. 

To what extent can we explain the behavior just using physical effort or just using physical risk? We correlated the predictions of the full model with participants' judgments of physical effort and physical risk and physical build times. We obtained confidence intervals on the model's correlation with the data by bootstrapping over participants in each dataset. We found that the fitted full model explained participants' judgments behavior better than either of effort or risk in each dataset. Interestingly, whereas in both the Estimated Time and Build Time dataset the effort-only model better explained behavior than the risk-only model ($p<0.05$ direct bootstrap hypothesis testing), in the Estimated Difficulty dataset the risk-only model performed significantly better than the effort-only model ($p<0.05$ direct bootstrap hypothesis testing; Figure~\ref{fig:allmodels}). To focus on the two judgment datasets, at the surface level, the responses of the Estimate Time and Estimated Difficulty participants are quite similar, and the full model indeed correlates with each of them to a similar degree. However, their responses appear to differentially correlate with the two separate components of our model: the risk-only model correlates better with the Estimated Difficulty dataset whereas the effort-only model correlates better with the Estimated Time dataset (in both comparisons $p<0.05$, direct bootstrap hypothesis testing; Figure~\ref{fig:allmodels}). This suggests that participants in these two groups might have attended to different features or components of the reconfiguration problems. Participants in the Estimated Difficulty experiment might have attended more to how difficult it would be to keep the structure stable. Participants in the Estimated Time experiment, on the other hand, seemed to have weighted the overall distance that each block would have to be moved to get from A to B in each pair, which is captured by physical effort in our model.

%% file: discussion.tex
\section{Discussion}

Intuitions about the difficulty of tasks are pervasive in our everyday decisions and actions. Yet, the representations and inferential processes that underlie these intuitions remain poorly understood. The current work brings together ideas and tools from cognitive science and robotics to formalize human adults' judgments and estimates of difficulty in precise, quantitative terms. Here we focused on people's intuitions about the difficulty of simple block-building tasks. We developed a novel computational model that quantifies the degree of difficulty to transform an initial state of blocks into a final target configuration. In doing so, our model takes into account not only the kinetic energy required to move blocks but also the uncertainty in the stability of the structure introduced by noisy placements of blocks. This model captures human judgments well by considering physical effort and the level of care required to achieve one's goals in the physical world.

Even though both cost components were important to account for the data, the extent to which participants considered each cost varied between tasks. We find that physical risk (i.e. the level of care required) was particularly important for people's estimates of task \textit{difficulty}, whereas considerations of physical effort strongly affected people's estimates about how much \textit{time} it would take to go from A to B.

The present study builds on \cite{yildirim2017physical}. The models presented in both studies use the same solver for path-planning \citep{toussaint2015logic} to accomplish trajectory optimization in multi-step complex action sequences. However, the present study differs from the previous work in two important ways. First, we account for uncertainty arising from the specific way in which the blocks might be scattered in the initial configuration. We also handle the mapping problem between states A and B (i.e., a particular wooden block in state A can go to one of many wooden blocks in state B) by choosing the assignment that minimizes the total Euclidean distance between the assigned pairs. 
Second, our model treats the contributions of physical risk (stability) and physical effort independently from each other whereas  \cite{yildirim2017physical} integrated stability tests within the intermediate stages of the geometric solver to filter out physically impossible plans. Our choice of isolating risk from effort was not only computationally more efficient 
but also allowed us to better understand human judgments across different tasks. 

The way in which our model evaluates and uses physical risks to plan suggests a novel approach to functionally characterize aspects of mental effort \citep{kool2018mental}. For example, when stacking blocks into a precarious configuration, we need to continuously attend to the task at hand, and finely plan our motor actions. More generally, our model emphasizes the role of internal object representations (e.g., shapes, positions, support relations, mass) as well internal body models that can support thinking about complex object manipulations. We see this intersection of physical reasoning, action planning, and difficulty estimation as a promising platform to formalize and reverse-engineer what computations are involved in the exertion of mental effort.

Interestingly, the model predictions showed relatively worse fits with how long it actually took participants to solve the different tasks compared to the estimated difficulty and time judgments. It's possible that people care about physical risk to different degrees at different points in the building process -- start more liberally but become more constrained and focused at the end (to avoid an almost completed tower from crashing).  We can model a change in the level of care depending on the current task at hand by having the planner reason about the physical risk at each time step and then actively controlling the level of noise $sigma$ as required. 

Our current model assesses stability only once at the end, without evaluating it in the intermediate stages throughout the execution of a plan. Arguably this is different from how a human planner would perform the task. While one might consider this as a limitation of the model, for this particular study, constraining the frequency of physical simulation was not only computationally convenient but also informative: it  allowed us to straightforwardly determine the contributions of effort and risk to explaining human judgments. Future work should consider adaptive allocation of computation to estimating stability throughout the intermediate stages of a plan as described above.

In current work, the level of noise in physics simulations was fit to the data. However, it would be interesting to try and infer the level of noise directly from people's actions. Different people will differ in how carefully they execute their actions, and in how skillful they are in interacting with the physical world. Recent work in cognitive development suggests that humans have representations of their own, as well as others' competence even early in life \citep{jara-ettinger2016naive}. These representations of competence can inform decisions about how to efficiently allocate tasks across agents \citep{magid2018four}, choose whom to help \citep{bennett2018preschoolers}, learn from others' effort \citep{leonard2017infants}, and even teach what would be too difficult for others \citep{bridgers2016children}. 

This study considered simple blocks-world type scenarios as a ``model organism'' to characterize the computations that support humans' intuitively inferences about task difficulty. One question is to what extent our current model can capture human judgments across different tasks or scenarios outside this domain. Although much further work is needed to capture people's sense of difficulty in non-physics domains (e.g., navigating an unfamiliar environment, solving a math problem, learning a new skill), we believe that a range of tasks that involve manipulation of physical objects could be modeled using our approach by adjusting a few aspects of the symbolic and geometric planner. Future work should explore how our approach can be extended to different manipulation tasks and to scenarios with more complex physics including objects with different shapes, density, and substance properties (e.g., soft bodies, liquids).

We look forward to these lines of future work that will help us better understand the inferential processes that underlie these behaviors, and build better artificial agents that can make these judgments in ways that humans do.